\documentclass{article}

\usepackage[preprint,nonatbib]{neurips_custom_2019}

\usepackage[utf8]{inputenc} 
\usepackage[T1]{fontenc}    
\usepackage{hyperref}       
\usepackage{url}            
\usepackage{booktabs}       
\usepackage{amsfonts}       
\usepackage{nicefrac}       
\usepackage{microtype}      
\usepackage{subfigure}
\usepackage{graphicx}
\usepackage{capt-of}
\usepackage{lineno}
\usepackage[]{authblk}
\usepackage{longtable}
\usepackage{tabularx}
\usepackage{multirow}

\newcolumntype{Y}{>{\centering\arraybackslash}X}

\usepackage{caption} 
\newcommand{\beginsupplement}{%
        \setcounter{table}{0}
        \renewcommand{\thetable}{S\arabic{table}}%
        \setcounter{figure}{0}
        \renewcommand{\thefigure}{S\arabic{figure}}%
     }

\graphicspath{{figures/}} 
\newcommand{\mycaption}[2]{\caption[#1]{\textbf{#1.} #2}}

\title{Machine Learning for Scent: Learning Generalizable Perceptual Representations of Small Molecules}
\author[1*]{Benjamin Sanchez-Lengeling}
\author[1*]{Jennifer N Wei}
\author[1]{Brian K Lee}
\author[2]{Richard C Gerkin}
\author[3]{Alán Aspuru-Guzik}
\author[1,$\dag$]{Alexander B Wiltschko}
\affil[1]{Google Research, Brain Team}
\affil[2]{School of Life Sciences, Arizona State University}
\affil[3]{Department of Chemistry, University of Toronto}
\affil[3]{Department of Computer Science, University of Toronto}
\affil[3]{Vector Institute for Artificial Intelligence, Toronto, Ontario, Canada}
\affil[3]{Canadian Institute for Advanced Research, Toronto, Ontario, Canada}
\affil[*]{Contributed equally}
\affil[$\dag$]{Email: alexbw@google.com}

\begin{document}

\maketitle

\begin{abstract}
Predicting the relationship between a molecule's structure and its odor remains a difficult, decades-old task. This problem, termed quantitative structure-odor relationship (QSOR) modeling, is an important challenge in chemistry, impacting human nutrition, manufacture of synthetic fragrance, the environment, and sensory neuroscience. We propose the use of graph neural networks for QSOR, and show they  significantly outperform prior methods on a novel data set labeled by olfactory experts. Additional analysis shows that the learned embeddings from graph neural networks capture a meaningful odor space representation of the underlying relationship between structure and odor, as demonstrated by a strong performance on two challenging transfer learning tasks. Machine learning has already had a large impact on the senses of sight and sound. Based on these early results with graph neural networks for molecular properties, we hope machine learning can eventually do for olfaction what it has already done for vision and hearing.
\end{abstract}

\section{Introduction}
Predicting properties of molecules is an area of growing research in machine learning \cite{Wu2018-yg, Mitchell2014-aw}, particularly as models for learning from graph-valued inputs improve in sophistication and robustness \cite{Scarselli2009-nn,Gilmer2017-oq}. A molecular property prediction problem that has received comparatively little attention during this surge in research activity is building Quantitative Structure-Odor Relationships (QSOR) models (as opposed to Quantitative Structure-Activity Relationships, a term from medicinal chemistry). This is a 70+ year-old problem straddling chemistry, physics, neuroscience, and machine learning \cite{Rossiter1996-nc}. 

Odor perception in humans is the result of the activation of 300-400 different types of olfactory receptors (ORs), expressed in millions of olfactory sensory neurons (OSNs), embedded in a small 5 $\textnormal{cm}^2$ patch of tissue called the olfactory epithelium. These OSNs send signals to the olfactory bulb, and then to further structures in the brain \cite{Su2009-bv, McGann2017-ir}. Advances in deep learning for vision and audition suggest that we might be able to directly predict the end sensory result of an input stimulus. Progress in deep learning for olfaction would aid in the discovery of new synthetic odorants, thereby reducing the ecological impact of harvesting natural products. Additionally, new representations of molecules derived from a model trained on odor recognition tasks may contribute our understanding of sensory perception in the brain \cite{Yamins2016-nf}.

Here, we curated a dataset of molecules associated with expert-labeled odor descriptors (in QSOR, \textit{odor descriptors} refer to the properties we wish to predict, as opposed to their usage in chemoinformatics, where they refer to the input features of a model). We trained Graph Neural Networks (GNNs) \cite{Gilmer2017-oq,Duvenaud2015-ye} to predict these odor descriptors using a molecule's graph structure alone. We show that our model learned a representation of odor space that clusters molecules based on perceptual similarity rather than purely on structural similarity, on both a global and local scale.
Further, we show that this representation is useful for making predictions on related tasks, which is a developing area in chemistry applications of machine learning \cite{Altae-Tran2017-yz, Fare2018-br}. These results indicate that our modeling approach has captured a general-purpose representation of the relationship between a molecule's structure and odor, which we anticipate to be useful for rational molecular design and screening.

\section{Prior Work in QSOR: A Decades-Long Pursuit}
\begin{figure}
  \centering
  \includegraphics[width=\textwidth]{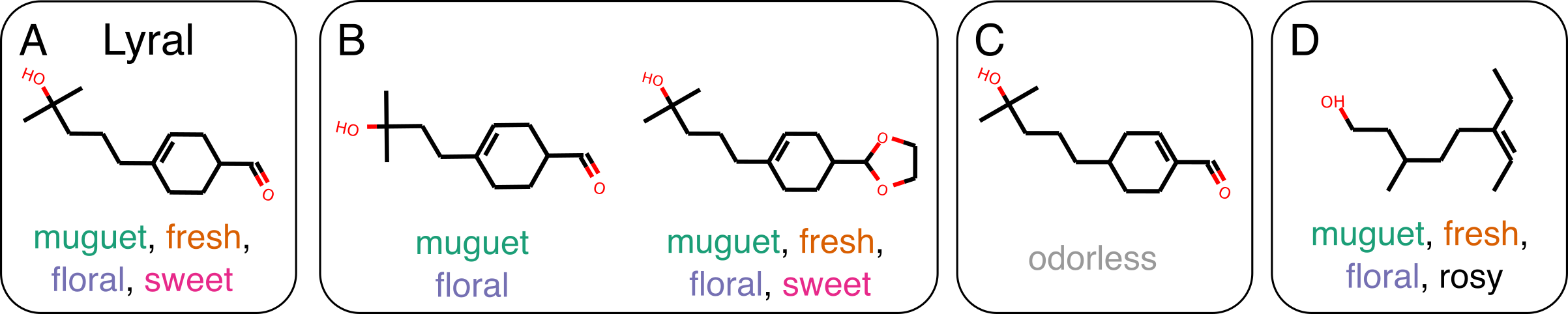}
  \setlength{\belowcaptionskip}{10pt} 
  \mycaption{Structurally similar molecules do not necessarily have similar odor descriptors}{\textbf{A.} Lyral, the reference molecule. \textbf{B.} Molecules with similar structure can share similar odor descriptors. \textbf{C.} However, a small structural change can render the molecule odorless. \textbf{D.} Further, large structural changes can leave the odor of the molecule largely unchanged. Example from Ohloff, Pickenhagen and Kraft \cite{Ohloff2012-bg}. \label{fig:lyral_example}}
\end{figure}

The problem of QSOR is ancient \cite{Sell2019-ti}, but in the scientific literature emerges with Amoore, Schiffman and Dyson, among others \cite{Amoore1964-ca, Dyson1937-ym, schiffman1974physicochemical}. Modern attempts to solve this problem in a directly data-driven and statistical manner began a few decades ago \cite{Rossiter1996-nc}, and even included early applications of neural networks  \cite{Chastrette1996-rm}. However, the number of odor descriptors used in these early studies was small (less than ten, usually one), and the number of total stimuli was limited (usually 10s, rarely 100s of molecules)~\cite{Sigma-Aldrich_Corporation2011-yk,Dravnieks1985-ek,Arctander1969-uf}. This has remained an open problem for so long due to its difficulty---very small changes in a molecule's structure can have dramatic effects on its odor, a phenomenon known in medicinal chemistry as an \textit{activity cliff} \cite{Sell2006-ag, Stumpfe2012-dc}. A classic example is \textit{Lyral}, which is a commercially successful molecule that smells of \textit{muguet} (a floral scent often used in dryer sheets). Its structural neighbors are not always perceptual neighbors, and some of its perceptual neighbors share little structural similarity (Figure \ref{fig:lyral_example}). 

Recently, the DREAM Olfactory Challenge spurred applications of traditional machine learning approaches to QSOR prediction \cite{Keller2017-vs}. This challenge presented a dataset where 49 untrained panelists rated 476 molecules on 21 odor attributes on an analog scale. The winning models of the DREAM challenge primarily relied on either the Dragon molecular features \cite{Mauri2006-sp} or Morgan fingerprints \cite{Rogers2010-uj} as a featurization of molecules. These features were used by random forests to make predictions, an approach with a long track record of success in chemoinformatics. We use these methods as baselines in this work.

We wish to highlight a few modern machine learning approaches to QSOR. Tran and colleagues \cite{Tran2018-nn} have revisited the use of neural networks for this task and have developed a convolutional neural network taking as input a custom 3D spatial representation of molecules. Nozaki et al. \cite{Nozaki2018-yg} used the mass spectra of molecules and natural language processing tools to predict textual descriptions of odor. Gutierrez et al. \cite{Gutierrez2018-ob} used word embeddings and chemoinformatics representations of molecules to predict odor properties.


\subsection{Classic Approaches to Featurizing Molecules and Modeling Their Properties}
QSOR has historically used many computational techniques from chemoinformatics and medicinal chemistry. For predicting molecular properties, molecules are typically transformed into fixed-length vectors using hand-crafted features, and fed to a prediction model such as a random forest or fully-connected neural network \cite{Mitchell2014-aw,Svetnik2003-ht}. We describe the details of baseline approaches to featurizing molecules below.

\subsubsection{Dragon and Mordred Features}
There are several available hand-crafted featurizations for molecules, which are popular in the field of olfactory neuroscience. Both \textit{Dragon} (closed source, \cite{Mauri2006-sp}) and \textit{Mordred} (open source, \cite{Moriwaki2018-hv}) are approaches that include many thousands of computed molecular features. They are an agglomeration of several types of molecular information and statistics, such as counts of atom types, graph topology statistics, and acid/base counts. Some of these features are easily interpretable (e.g. \textit{number of Carbon atoms}) and some are not (e.g. \textit{spectral moment of order 4 from distance/detour matrix}). We use \textit{Mordred} in the present work because it is open source, and we found no appreciable difference in predictive performance between these features and \textit{Dragon} features (data not shown). 

\subsubsection{Molecular Fingerprints}

\textit{Molecular fingerprints} encode topological environments of a molecular graph into a fixed-length vector. An environment is a fragment of the molecular graph, and indicates the presence of a single atom type or a functional group, e.g. an alcohol or ester group. This approach to featurizing molecules is popular in the field of medicinal chemistry; traditionally, bit-based Morgan fingerprints have been used in chemoinformatics for retrieving nearest neighbor molecules using Tanimoto similarity \cite{Maggiora2014-va}. When these environments are atom-centered and constructed via adjacent atoms, they are called Extended-Connectivity Fingerprints, or \textit{Morgan fingerprints} \cite{Rogers2010-wo} and when they are constructed via paths through the graph they are \textit{path descriptor fingerprints} \cite{Randic1999-fn}. The more commonly used bit variant records the presence of a given environment (e.g., is there an ester in this molecule?), while the count variant records the number of instances of a given environment (e.g. how many ester groups are there in this molecule?). This information is hashed into a fixed-length vector. There are two tunable parameters: max topological radius and fingerprint vector size. The max topological radius determines the largest fragment which the fingerprint can represent. Fingerprint vector size affects how likely a hash collision can occur. We tune both of these parameters to maximize predictive performance.

In our baseline experiments, we explicitly compare bit-based path descriptors fingerprints (bFP) and count-based Morgan fingerprints (cFP). The cheminformatics package RDKit was used to generate both types of fingerprints \cite{rdkit}. Molecular properties are typically predicted using models such as random forests or support vector machines, so we use random forests as the predictive model for each of the bFP and cFP features.

\section{Graph Neural Networks}

Most machine learning models require regularly-shaped input (e.g. a grid of pixels, or a vector of numbers) as input. Recently, Graph Neural Networks (GNNs) have enabled the use of irregularly-shaped inputs, such as graphs, to be used directly in machine learning applications \cite{Wu2019}. Fields of use include predicting friendships in social network graphs, citation networks in academic literature, and most germane for this work, classification and regression tasks in chemistry \cite{Wu2018-yg}. 

\subsection{Graph Neural Networks for Predicting Molecular Properties}

By viewing atoms as nodes, and bonds as edges, we can interpret a molecule as a graph. GNNs are learnable permutation-invariant transformations on nodes and edges, which produce fixed-length vectors that are further processed by a fully-connected neural network. GNNs can be considered learnable featurizers specialized to a task, in contrast with expert-crafted general features \cite{Gilmer2017-oq,Duvenaud2015-ye}. GNNs have achieved state-of-the-art results in the prediction of biophysical, biological, physical, and electronic quantum properties of molecules \cite{Wu2018-yg}, and thus, we believe their use in QSOR to be promising.

\begin{figure}[h]
  \centering
  \includegraphics[width=\textwidth]{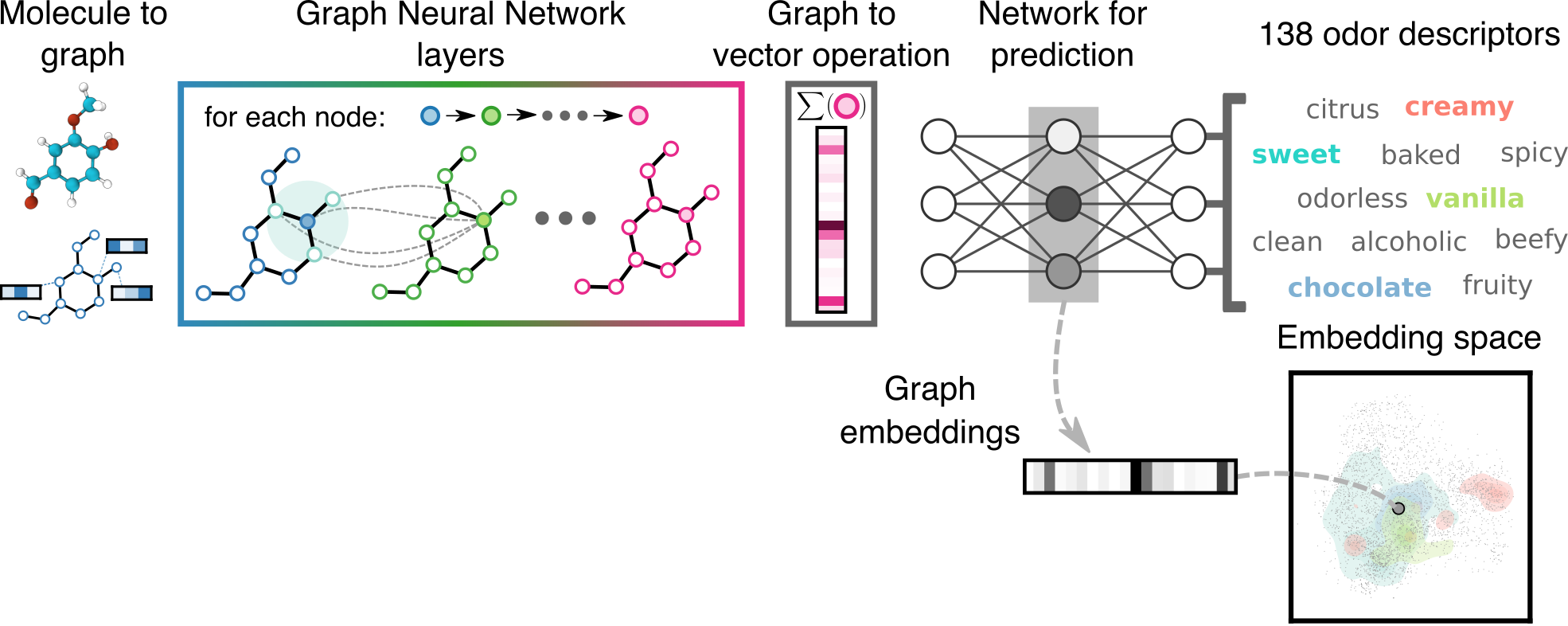}
  \mycaption{Model Schematic}{
  Each molecule is first featurized by its constituent atoms, bonds, and connectivities. Each Graph Neural Network (GNN) layer, here represented as different colors, transforms the features from the previous layer. The outputs from the final GNN layer is reduced to a vector, which is then used for predicting odor descriptors via a fully-connected neural network. We retrieve graph embeddings from the penultimate layer of the model. An example of the embedding space representation for four odor descriptors is shown in the bottom right; the colors of the regions in this plot correspond to the colors of odor descriptors in top right. \label{fig:model} }
\end{figure}

The GNN consists of message passing layers, each followed by a reduce-sum operation, followed by several fully connected layers.  Architectural details can be found in the the appendix, Table \ref{si:hyper}. The final fully-connected layer has a number of outputs equal to the number of odor descriptors being predicted. Figure \ref{fig:model} illustrates our model. We implement these GNN models using the TensorFlow software package \cite{abadi2016tensorflow}.

\subsection{Learned Graph Neural Network Embeddings}
All deep neural network architectures build representations of input data at their intermediate layers. The success of deep neural networks in prediction tasks relies on the quality of their learned representations, often referred to as embeddings \cite{Bengio2013-gf}. For instance, ImageNet embeddings are often used as-is to make predictions on unrelated image tasks \cite{donahue2014decaf, sharif2014cnn}, and with the advent of the BERT model and its cousins, this ability to use pre-trained embeddings is becoming common in natural language processing \cite{Devlin2018-rp}. The structure of a learned embedding can even lead to insights on the task or problem area, and the embedding can even be an object of study itself \cite{Yamins2016-nf, coenen2019visualizing}.

We save the activations of the penultimate fully connected layer as a fixed-dimension “odor embedding”. The GNN model must transform a molecule's graph structure into a fixed-length representation that is useful for classification. Although the utility of learned neural network embeddings of molecules is still young and relatively unproven \cite{Gomez-Bombarelli2018-ar,Zhavoronkov2019-pw}, we still anticipate that a learned GNN embedding on an odor prediction task may include a semantically meaningful and useful organization of odorant molecules. We explicitly test the utility of this odor embedding in later sections in this work.

\section{A Curated QSOR Dataset}
\begin{figure}[h]
  \centering

  \includegraphics[width=\textwidth]{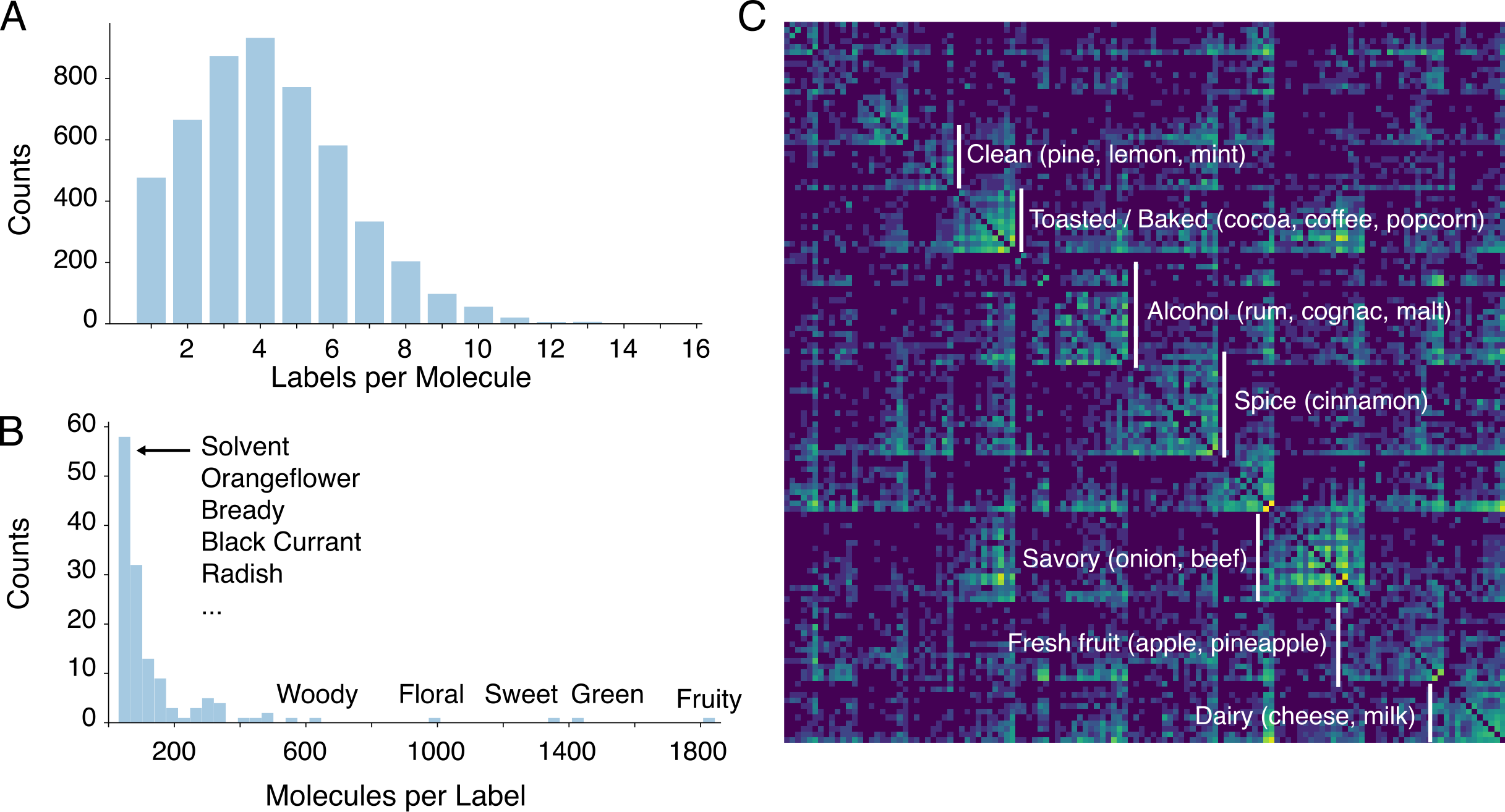}
  \mycaption{Dataset overview}{ \textbf{A.} Distribution of odor descriptor frequencies. \textbf{B.} Distribution of label density. \textbf{C.} Co-ocurrence matrix for odor descriptors. The 10 most frequent descriptors are removed for visual clarity, and remaining descriptors re-ordered using spectral clustering. Main odor groups with examples are highlighted. The color range is on a log-scale, and normalized such that each row and column sums to 1. \label{fig:dataset} }
\end{figure}

We assembled an expert-labeled set of 5030 molecules from two separate sources: the \textit{GoodScents} perfume materials database ($n=3786$, \cite{noauthor_undated-uu}) and the \textit{Leffingwell PMP 2001} database ($n=3561$, \cite{Leffingwell2005-uv}). The datasets share $2317$ overlapping molecules. Molecules are labeled with one or more odor descriptors by olfactory experts (usually a practicing perfumer), creating a multi-label prediction problem. GoodScents describes a list of 1--15 odor descriptors for each molecule (Figure \ref{fig:dataset}A), whereas Leffingwell uses free-form text. Odor descriptors were canonicalized using the GoodScents ontology, and overlapping molecules inherited the union of both datasets' odor descriptors. After filtering for odor descriptors with at least 30 representative molecules, 138 odor descriptors remained (Figure \ref{fig:dataset}B), including an \textit{odorless} descriptor. Some odor descriptors were extremely common, like \textit{fruity} or \textit{green}, while others were rare, like \textit{radish} or \textit{bready}. This dataset is composed of materials for perfumery, and so is biased away from malodorous compounds. There is also skew in label counts resulting from different levels of specificity, e.g. \textit{fruity} will always be more common than \textit{pineapple}. 

There is an extremely strong co-occurrence structure among odor descriptors that reflects a common-sense intuition of which odor descriptors are similar and dissimilar (Figure \ref{fig:dataset}C). For example, there is a \textit{dairy} cluster that includes the \textit{dairy}, \textit{yogurt}, \textit{milk}, and \textit{cheese} descriptors, indicating that they often co-occur as descriptors in individual molecules. There is also a \textit{fruity} cluster with \textit{apple}, \textit{pear}, \textit{pineapple} etc., and a \textit{bakery} cluster that includes \textit{toasted}, \textit{nutty}, and \textit{cocoa}, among others. Previous approaches in QSOR often train one model per odor descriptor. To take advantage of this correlation structure, we apply a GNN to predict all 138 odor descriptor tasks at once.

\section{QSOR Prediction Performance Benchmark}

\begin{figure}
    \centering
    \includegraphics[width=0.75\columnwidth]{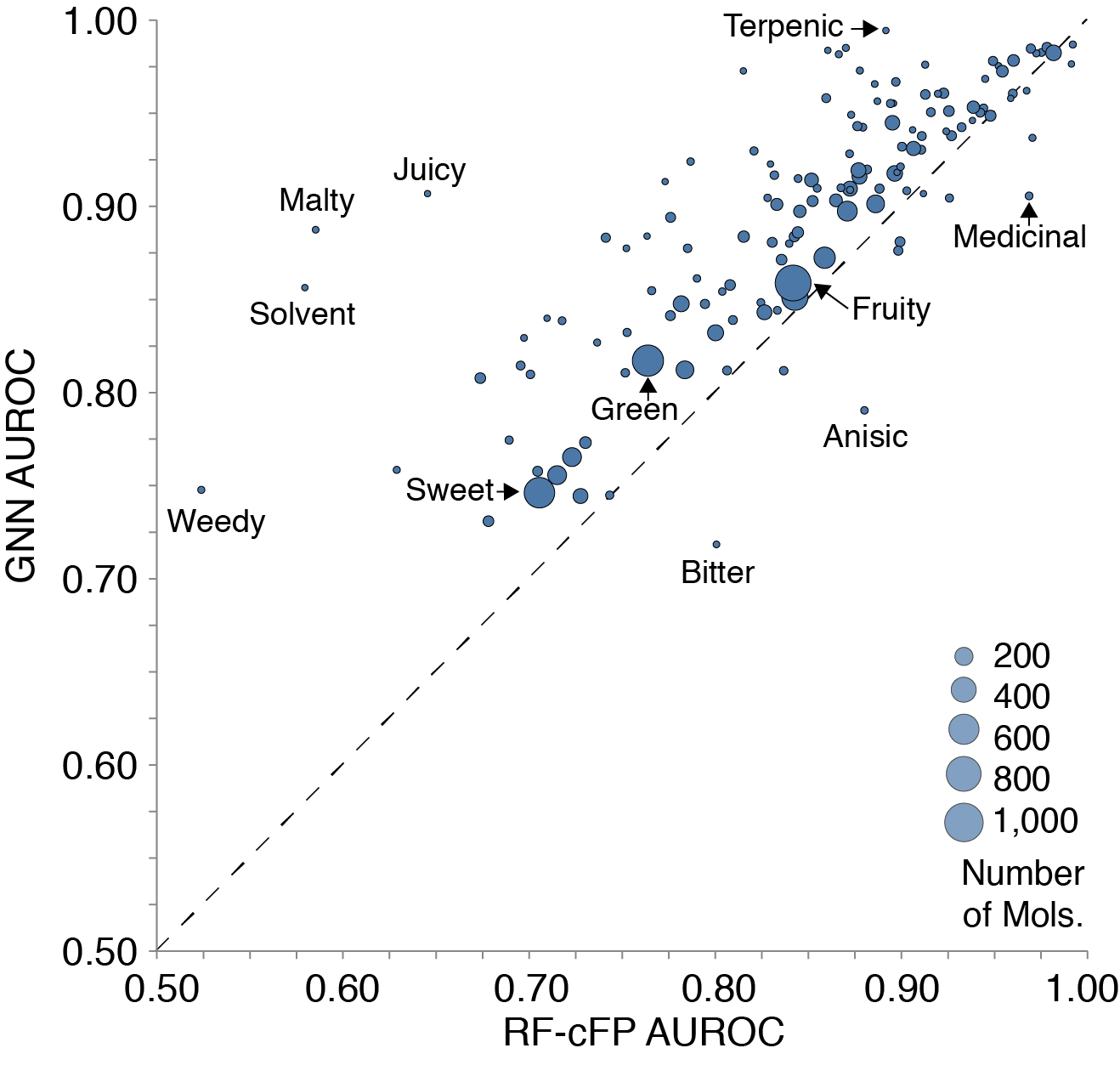}
    \mycaption{Comparison of RF-cFP and GNN, broken down by odor descriptor}{Each dot represents an odor descriptor, with size representing the number of positive examples. GNN outperforms RF-cFP on nearly all odor descriptors. \label{fig:classification-performance}}
\end{figure}

We benchmark classification performance for each odor descriptor in our dataset, as a multi-label classification problem. We compare the GNN model against random forest models (RF) and k-nearest neighbor models (KNN) on bit-based RDKit fingerprints (bFP), count-based Morgan fingerprints (cFP), and Mordred features. We report several metrics (Table \ref{tab:classification-results}), as each metric can highlight different performance characteristics. For the rest of the analysis, we primarily compare models on mean AUROC, averaged across odor descriptors; AUROC performance by descriptor is shown in Figure \ref{fig:classification-performance}. We trained non-graph based fully-connected neural networks on cFP and bFP features, but their performance is indistinguishable from the RF model (data not shown).

\begin{table}[h]
    \centering
    \resizebox{\columnwidth}{!}{
\begin{tabular}{rccc}
    \toprule 
    & AUROC & Precision & F1 \\\midrule
    GNN & \textbf{0.894 [0.888, 0.902]} & \textbf{0.379 [0.351, 0.398]} & \textbf{0.360 [0.337, 0.372]} \\
    RF-Mordred & 0.850 [0.838, 0.860] & 0.311 [0.288, 0.333] & 0.306 [0.283, 0.319] \\
    RF-bFP & 0.832 [0.821, 0.842] & 0.321 [0.293, 0.339]  & 0.295 [0.272, 0.308] \\
    RF-cFP & 0.845 [0.835, 0.854] & 0.315 [0.280, 0.332] & 0.295 [0.272, 0.311] \\
    KNN-bFP & 0.791 [0.778, 0.803] & 0.328 [0.305, 0.347]  & 0.323 [0.299, 0.335] \\
    KNN-cFP & 0.796 [0.785, 0.809] & 0.333 [0.307, 0.351] & 0.316 [0.292, 0.327] \\
    \bottomrule
\end{tabular}
    }
    \mycaption{Odor descriptor prediction results}{mean, 95\% CI [lower, upper] bounds reported. Numbers reported are an unweighted mean across all 138 odor descriptors; see Supplemental Table \ref{tab:detailed-classification-results} for results reported by odor label. Precision/recall decision thresholds are optimized for F1 score on a cross-validation split created from the training set.  The best values for each metric are in bold. Models include graph neural networks (GNN), random forest (RF) and k-nearest neighbor. Featurizations include bit-based RDKit fingerprints (bFP), count-based Morgan fingerprints (cFP), and Mordred features.
    There was no statistical winner as measured by recall, and thus it is omitted; these scores ranged from 0.365 to 0.393, with high overlap amongst all models.\label{tab:classification-results}}
\end{table}






\section{Evaluating Odor Embeddings}

An odor embedding representation that reflects common-sense relationships between odors should show structure both globally and locally. Specifically, for global structure, odors that are perceptually similar should be nearby in an embedding. For local structure, individual molecules that have similar odor percepts should cluster together and thus be nearby in the embedding. We examine both of these properties in sequence.

\subsection{Examining the Global Structure of a Learned Odor Space}

We take our embedding representation of each data point from the penultimate-layer output of a trained GNN model. In the case of our best model, each molecule gets mapped to a 63-dimensional vector. 
Qualitatively, to visualize this space in 2D we use principal component analysis (PCA) to reduce its dimensionality. The distribution of all molecules sharing a similar label can be highlighted using kernel density estimation (KDE). 

\begin{figure}[h]
  \centering
  \includegraphics[width=\textwidth]{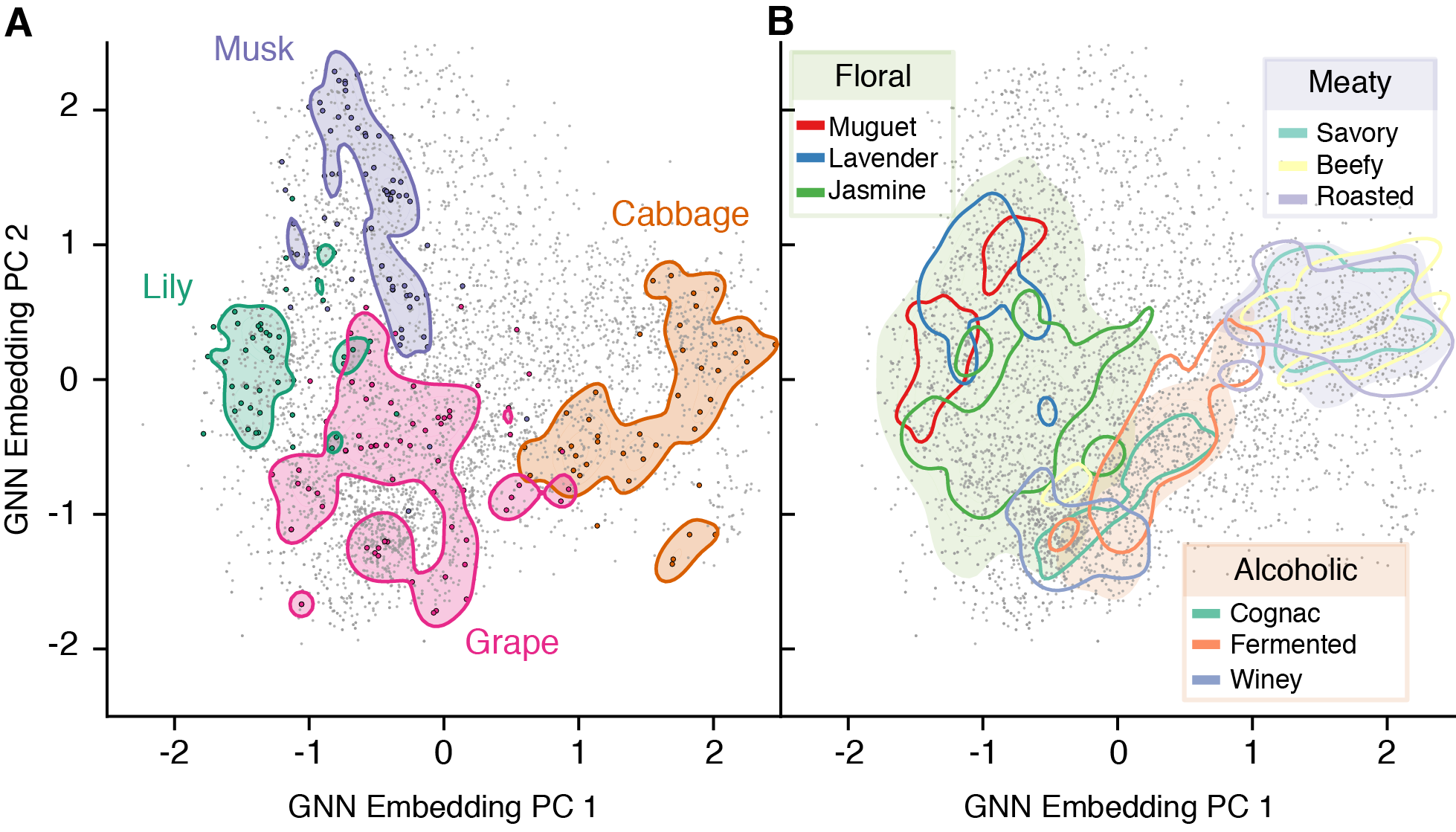}
  \mycaption{2D representation of a GNN model embeddings as a learned odor space}{Molecules are represented as individual points. Shaded and contoured areas are kernel density estimates of the distribution of labeled data. \textbf{A}. Four odor descriptors with low co-occurrence have low overlap in the embedding space. \textbf{B}. Three general odor descriptors (\textit{floral}, \textit{meaty}, \textit{alcoholic}) each largely subsume more specific labels within their boundaries. See Supplemental Figure \ref{fig:fp_space} for the equivalent analysis with molecular fingerprints.
  \label{fig:odorspace}}
\end{figure}

\begin{figure}[h]
  \centering
  \includegraphics[width=\textwidth]{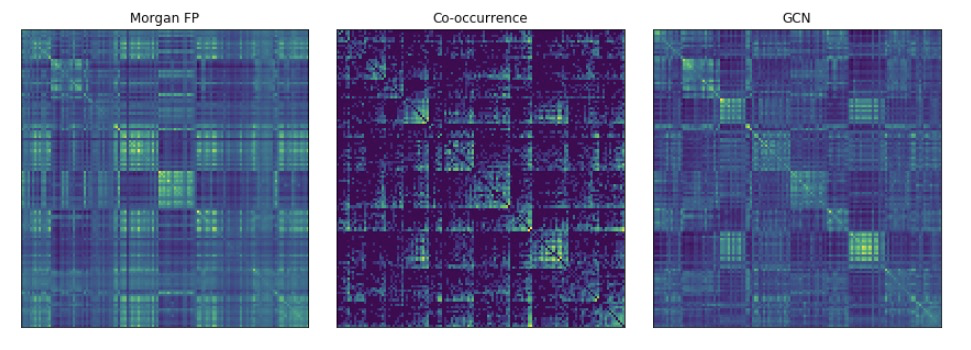}
  \mycaption{A comparison of embedding distances with odor label co-occurrence }{We compare the odor label co-occurrence (center) with the Morgan fingerprint embedding (left) and the GNN embedding (right). For the fingeprint embedding and the GNN embeddings, the cosine distance between the embeddings of two molecules sharing the corresponding label are depicted. The correlation coefficient between the GCN embedding image and the label co-occurrence matrix is 0.43, and the correlation coefficient between the fingerprint embedding and the label co-occurrence matrix is 0.22. The color range for each matrix is on a log-scale, and the matrix is normalized such that per-row and per-column sums equal 1.\label{fig:embedding_cooccurrence}} 
\end{figure}

\begin{figure}[h]
  \centering
  \includegraphics[width=\textwidth]{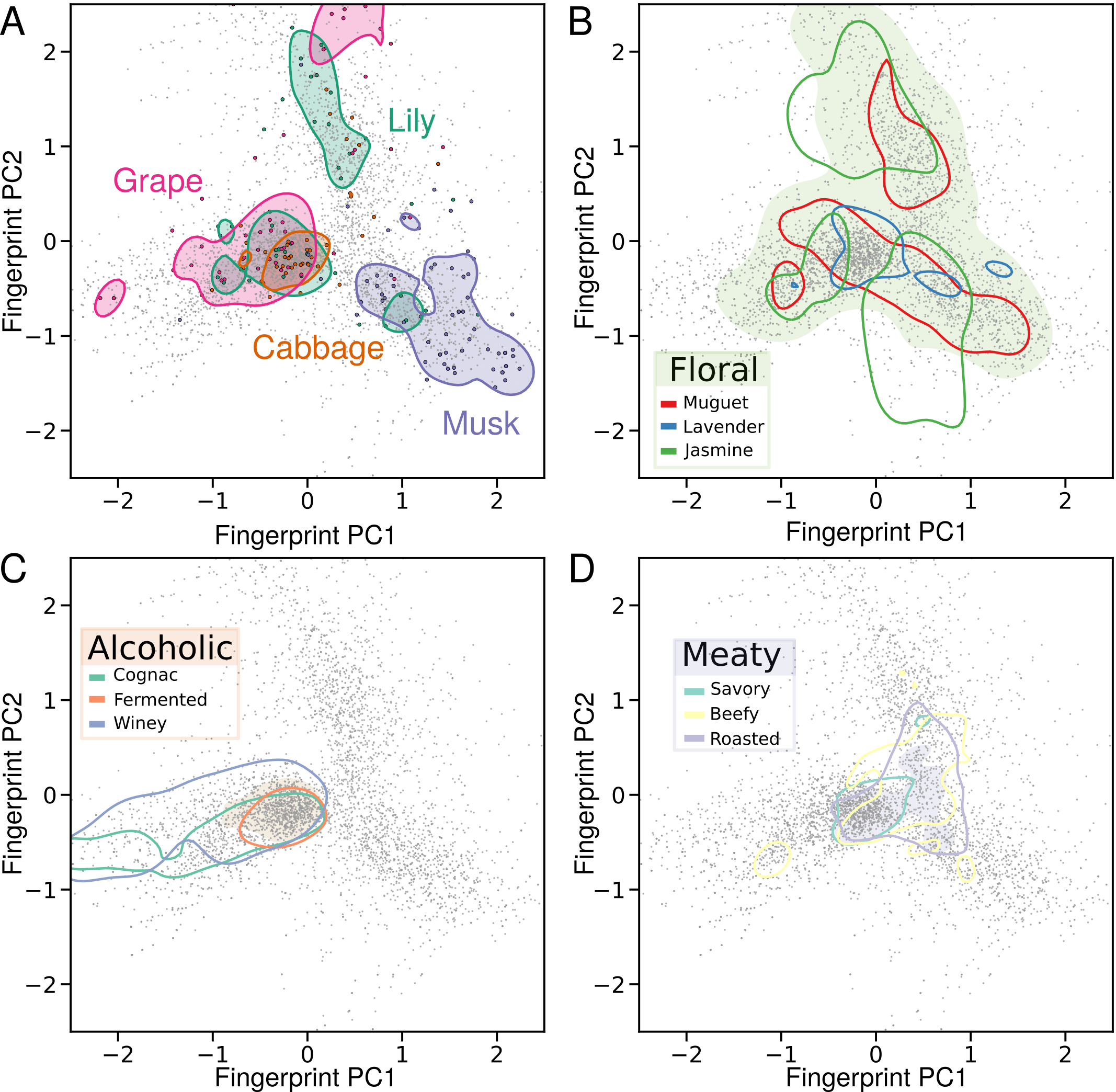}
  \mycaption{2D representation of molecular fingerprints as an unlearned odor space}{Fingerprints are dimensionally-reduced to two dimensions using PCA. For clarity, molecules are assigned a z-score based on a Gaussian fit of the data, and molecules with $z>2.5$ are not shown. Contoured and shaded areas are computed via KDE of positive labeled data identically to Figure \ref{fig:odorspace}. \textbf{A.} Labels with low co-occurrence are spread across the embedding space, but show substantial overlap, as opposed to the GNN-based embeddings. \textbf{B}, \textbf{C} and \textbf{D} each show an individual general label (\textit{Floral}, \textit{Alcoholic} and \textit{Meaty}), and three more specific versions that should be contained in each label. On the whole, the embedding space does not reflect the hierarchical organization of odor descriptors as reflected in both the co-occurrence matrix (Figure \ref{fig:dataset}C) or the GNN embeddings learned from the data (Figure \ref{fig:odorspace}).  \label{fig:fp_space}}
\end{figure}

The global structure of the embedding space is illustrated in Figure \ref{fig:odorspace}. In this example, we find that individual odor descriptors (e.g. \textit{musk}, \textit{cabbage}, \textit{lily} and \textit{grape}) tend to cluster in their own specific region. For odor descriptors that co-occur frequently, we find that the embedding space captures a hierarchical structure that is implicit in the odor descriptors. The clusters for odor labels \textit{jasmine}, \textit{lavender} and \textit{muguet} are found inside the cluster for the broader odor label \textit{floral}. If we examine the pairwise distances between all odors in our learned embedding, we see the block structure apparent in Figure \ref{fig:dataset}C is reflected by the learned GNN embedding, but not with molecular fingerprints (Figure \ref{fig:embedding_cooccurrence}). Further, a dimensionally-reduced molecular fingerprint does not share the same degree of organization and interpretability (Figure \ref{fig:fp_space}).


\subsection{Evaluating the Local Structure of a Learned Odor Space}

We tested whether molecules nearby in embedding space share perceptual similarity. Specifically, we asked whether molecules with small cosine distances in our GNN embeddings were perceptually similar. As a baseline, we used Tanimoto distance, which is equivalent to Jaccard distance on bFP features. Tanimoto distance is a commonly used metric for molecular database lookup in chemoinformatics. However, molecules with similar structural features do not always smell the same (Figure \ref{fig:lyral_example}), so we anticipated that nearest neighbors using bFP features may not be as perceptually similar as neighbors in using our embeddings. 

\begin{figure}[h]
  \centering
  \includegraphics[width=\textwidth]{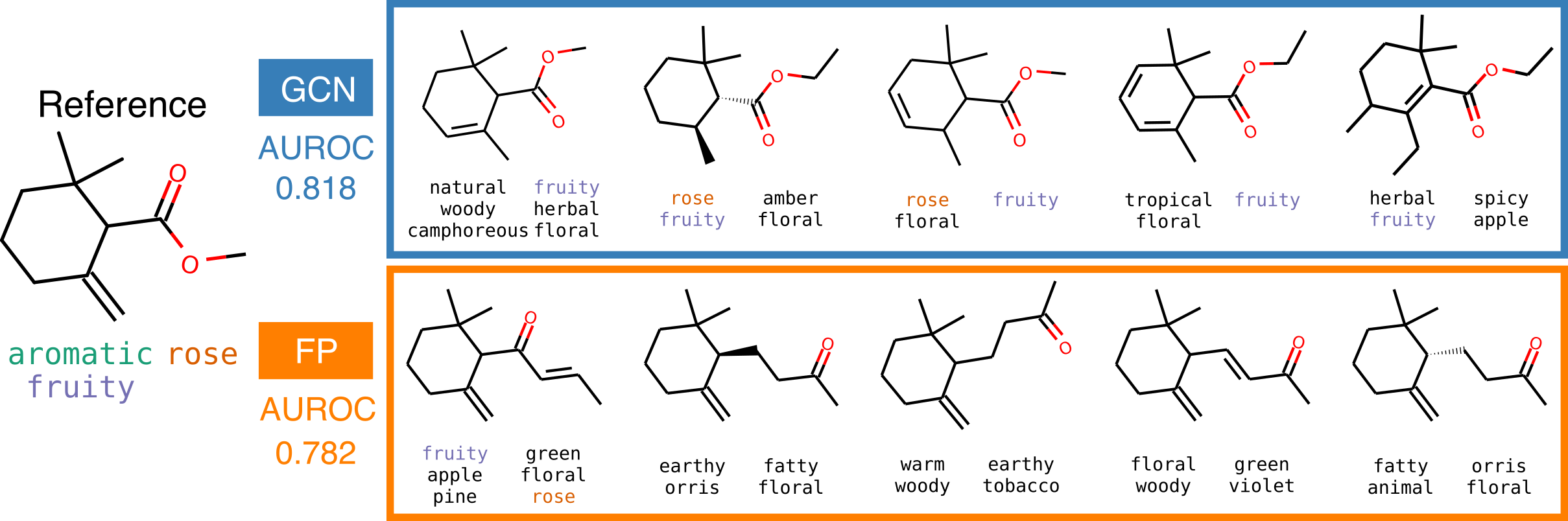}
  \mycaption{Nearest neighbor retrieval}{The top-five nearest neighbors to damascone carboxylate are shown for cosine similarity on GNN embeddings and for Tanimoto distance on bit-based Morgan fingerprints. The AUROCs are shown for k-Nearest Neighbor classifier ($k=20$) performance averaged over all odor descriptors, trained with the corresponding feature representation.   \label{fig:neighbor_lookup}}
\end{figure}

\begin{figure}[h]
  \centering
  \includegraphics[width=0.5\textwidth]{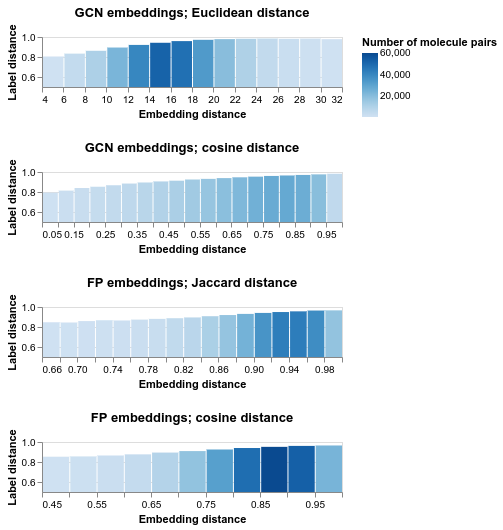}
  \mycaption{Label distance vs. embedding distance under different metrics}{
We found that label distance (as measured by Jaccard distance) correlated with distance in embedding space for different choices of embedding space and distance metrics. We calculated all pairwise distances between our training set and test set molecules, for the following spaces and metrics: GNN embeddings/Euclidean distance, GNN embeddings/cosine distance, Morgan bFP embeddings/Jaccard distance, Morgan bFP embeddings/cosine distance.}
\label{fig:label_embed_distance} 
\end{figure}

We trained a k-nearest neighbors (KNN) classifier ($k=20$) to predict odor descriptors from GNN embeddings and bFPs. 
GNN embeddings (AUROC = 0.818, 95\% CI [0.806, 0.830] ) outperformed bFP (AUROC = 0.782, 95\% CI [0.773, 0.797]). Inspecting the nearest neighbors found by each method (Figure \ref{fig:neighbor_lookup}) reveals that both methods yield molecules with similar structural features, but retrieval using GNN embeddings yields molecules that are more perceptually similar to the source molecule.
This suggests that our representations are better able to cluster molecules by their odor perceptual similarity than bit-based fingerprints. Figure \ref{fig:label_embed_distance} and Table \ref{tab:pairwise_distance_odor_label_table} show additional results comparing odor perceptual similarity and embedding distance between molecules using different distance metrics with bFPs and GNN embeddings.


We have shown that our embedding space has global and local structure that reflect the common-sense and psychophysical organization of odor descriptors. In the following sections, we show that this organization is \textit{useful}, and that this embedding can be used to make predictions on adjacent, challenging tasks.

\subsection{Transfer Learning to Previously-Unseen Odor Descriptors}

\begin{figure}[h]
\centering
  \includegraphics[width=0.5\textwidth]{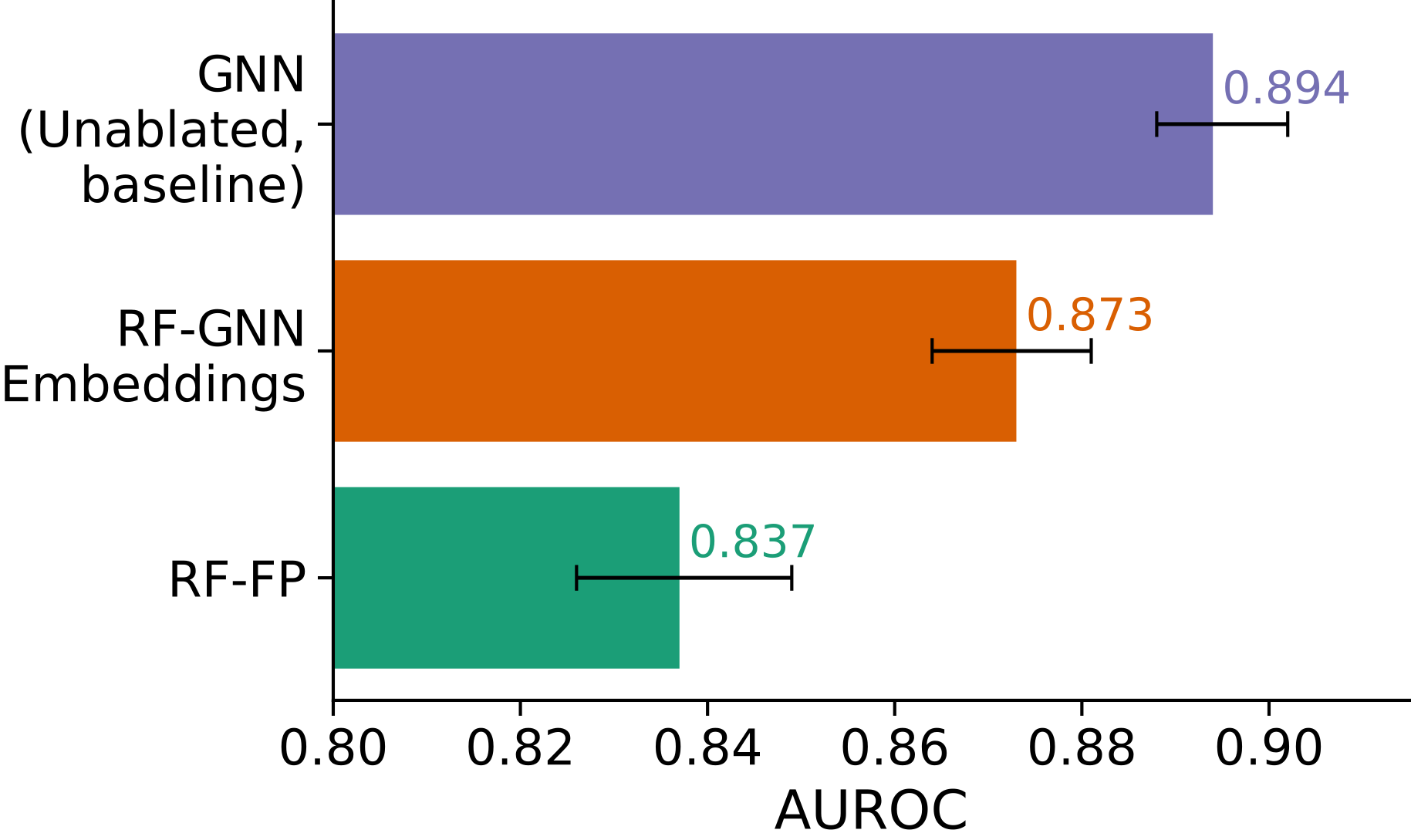}
  \mycaption{Mean AUROC on a previously held-out odor}{Average AUROC scores across all labels on the single label ablation task. The error bars denote 95\% confidence intervals. The top bar denotes the performance of the model trained on all of the labels. The middle bar denotes the performance of a random forest model trained using the GNN embeddings from a model trained on $(N-1)$ odor labels. The bottom bar denotes a random forest trained on counting Morgan fingerprints.  \label{fig:ablation_results}}
\end{figure}

An odor descriptor may be newly invented or refined (e.g., molecules with the \textit{pear} descriptor might be later attributed a more specific \textit{pear skin, pear stem, pear flesh, pear core} descriptor). A useful odor embedding would be able to perform transfer learning \cite{Pan2010-jk} to this new descriptor, using only limited data. 
To approximate this scenario, we ablated one odor descriptor at a time from our dataset. Using the embeddings trained from $(N-1)$ odor descriptors as a featurization, we trained a random forest to predict the previously held-out odor descriptor. We used cFP and Mordred features as a baseline for comparison. The results are shown in Figure \ref{fig:ablation_results}.  GNN embeddings significantly outperform Morgan fingerprints and Mordred features on this task, but as expected, still perform slightly worse than a GNN trained on the target odor. This indicates that GNN-based embeddings may generalize to predict new, but related, odors.

\subsection{Generalizing to Other Olfaction Tasks: the DREAM Olfaction Prediction Challenge}

\begin{figure}[!h]
  \centering
  \includegraphics[width=0.6\textwidth]{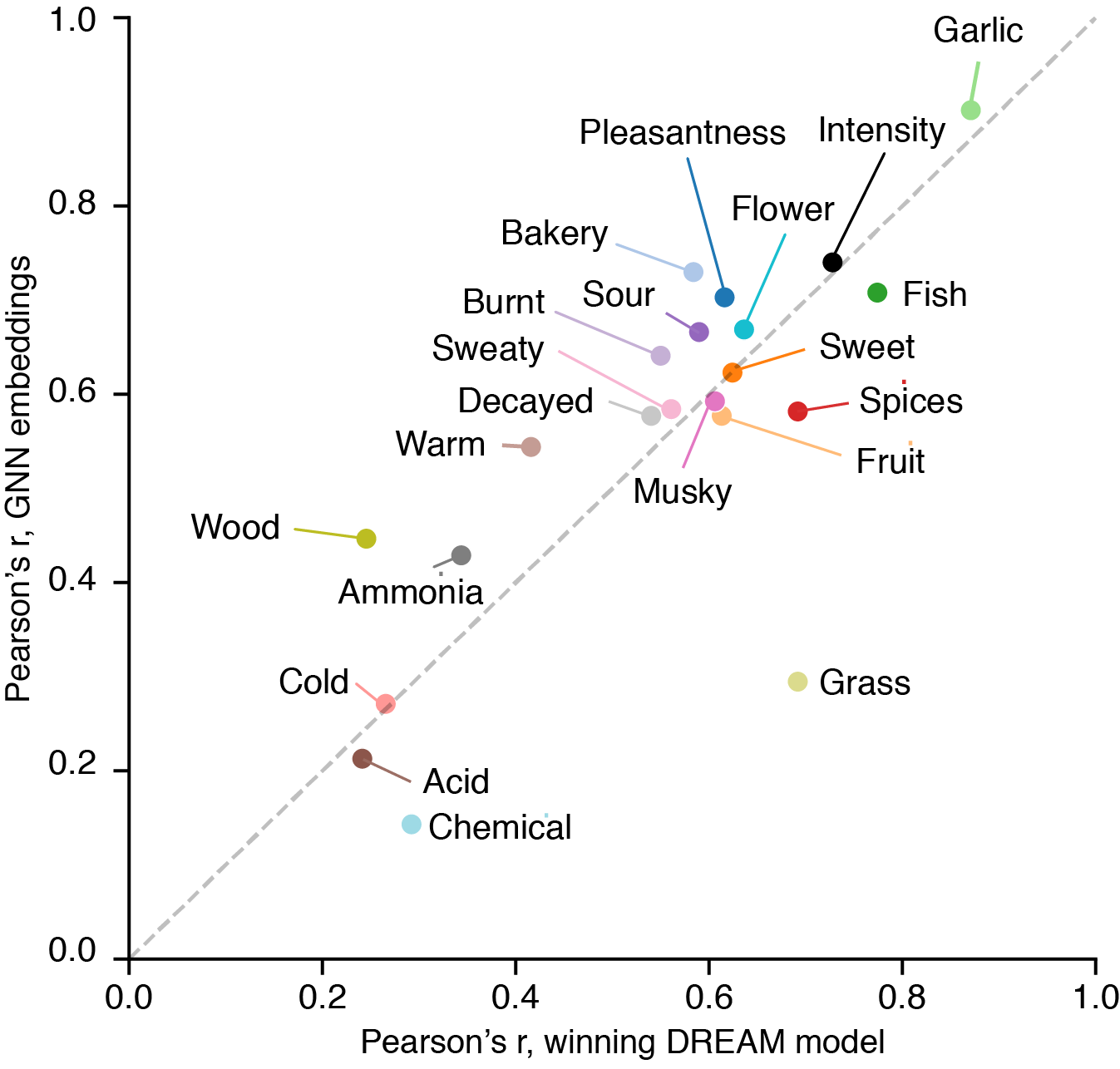}
  \mycaption{Predictive performance of GNN and best baseline model on the the DREAM Olfaction Prediction Challenge}{Pearson's $r$ for DREAM challenge winner (for most odors, a random forest on a subset of \textit{Dragon} features) versus a RF trained on GNN embeddings, broken down by odor descriptor. Dotted gray represents equal performance for both models. Data points above the diagonal line indicate better GNN predictive performance. Although mean values are generally higher for the GNN model, both the state-of-the-art model and GNN are statistically indistinguishable. \label{fig:dream_r}} 
\end{figure}

\begin{figure}[h]
  \centering
  \includegraphics[width=\textwidth]{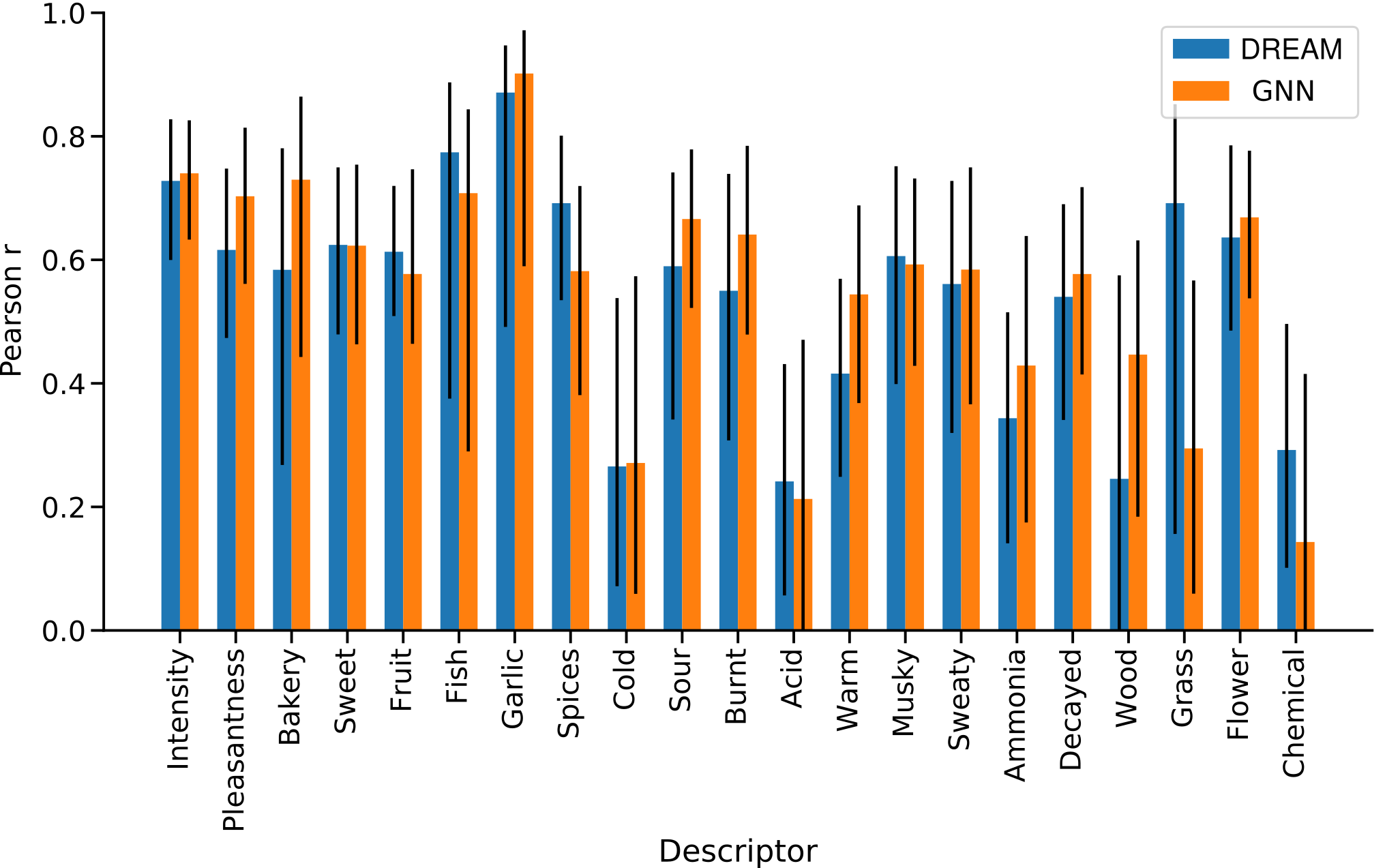}
  \mycaption{DREAM Pearson's r with confidence intervals}{
  Bar chart of Pearson's $r$ for DREAM challenge winner versus a RF trained on GNN embeddings, broken down by odor descriptor and approach. 95\% CI intervals are computed via bootstrap. The predictions transfer-learned using a random forest on GNN embeddings are statistically indistinguishable from the state-of-the-art DREAM model.
  \label{fig:dream_r_ci}}
\end{figure}

\begin{figure}[h]
  \centering
  \includegraphics[width=\textwidth]{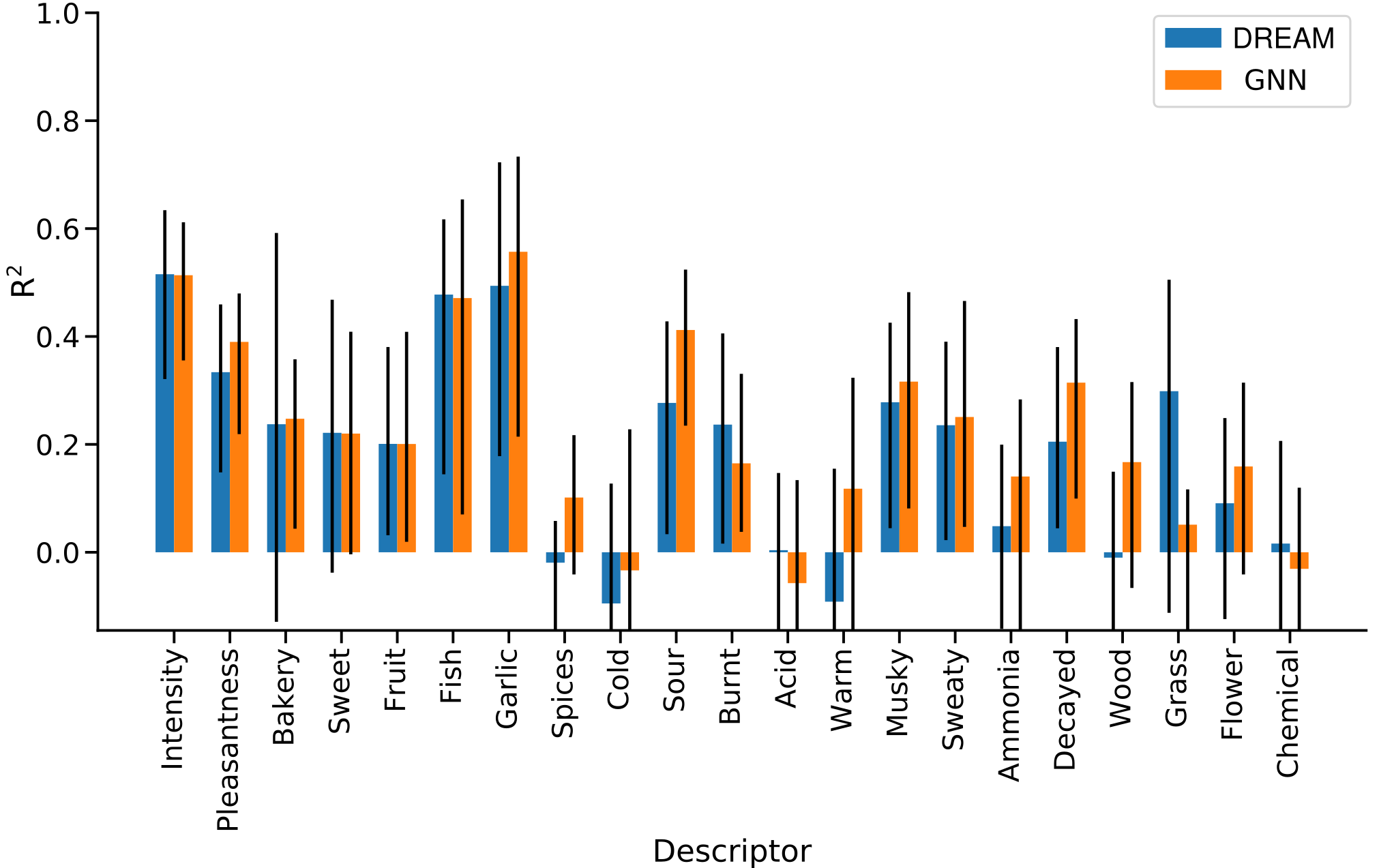}
  \mycaption{DREAM olfaction challenge $R^2$ with confidence intervals}{Bar chart of $R^2$ for DREAM challenge winner versus a RF trained on GNN embeddings, broken down by odor descriptor and approach. 95\% CI intervals are computed via bootstrap. The predictions transfer-learned using GNN embeddings are statistically indistinguishable from the state-of-the-art DREAM model. \label{fig:dream_R2_ci}}
\end{figure}

The DREAM Olfaction Prediction Challenge \cite{Keller2017-vs} was an open competition to build QSOR models on a dataset collected from untrained panelists. The DREAM dataset has several differences from our own. First, it was a regression problem —-- panelists rated the amount that a molecule smelled of a particular odor descriptor on a scale from 1 to 100. Second, it had 476 molecules compared to our $\sim5$k (although our dataset contains nearly all of the DREAM molecules). Third, the ratings were provided by a large panel of untrained individuals over a short period of time, whereas ours were gleaned from a small set of experts over many years. The DREAM challenge measured model performance as the Pearson's $r$ correlation of model predictions with the mean reported intensity of each odor descriptor, which we show in Figure \ref{fig:dream_r}. Additional statistics such as $R^2$ and 95\% confidence intervals are found in Figures \ref{fig:dream_r_ci}, \ref{fig:dream_R2_ci}.

The winning DREAM model used random forest models with a combination of several sources of features, primarily Dragon and Morgan fingerprints, among other sources of information \cite{Keller2017-vs}.
Using only our embedding with a tuned random forest model, we achieve a mean Pearson's $r = 0.55$ while the state-of-the-art model described above achieved a mean Pearson's $r = 0.54$. While we can have better average performance in 13 tasks, when taking into account confidence intervals, we find the performance is indistinguishable between the two models for both $r$ and $R^2$ regression scores (Figures \ref{fig:dream_r_ci}, \ref{fig:dream_R2_ci}).

Overall, this indicates that our QSOR modeling approach can generalize to adjacent perceptual tasks, and captures meaningful and useful structure about human olfactory perception, even when measured in different contexts, with different methodologies.

\section{Conclusion}

We assembled a novel and large dataset of expertly-labeled single-molecule odorants, and trained a graph neural network to predict the relationship between a molecule’s structure and its smell. We demonstrated state-of-the-art results on this QSOR task with respect to field-recognized baselines. Further, we showed that the embeddings capture meaningful structure on both a local and global scale. Finally, we showed that the embeddings learned by our model are useful in downstream tasks, which is currently a rare property of modern machine learning models and data in chemistry. Thus, we believe our model and its learned embeddings might be generally useful in the rational design of new odorants.

\section{Acknowledgments}

We thank D. Sculley, Steven Kearnes, Jasper Snoek, Emily Reif, Carey Radebaugh, David Belanger, Joel Mainland and Emily Mayhew for support, suggestions and useful discussions on the manuscript. We thank Max Bileschi and Yoni Halpern for their technical guidance. We thank Aniket Zinzuwadia for discussion and who did preliminary work on the DREAM and GoodScents datasets for his senior thesis. We thank Bill Luebke of GoodScents and John Leffingwell of Leffingwell and Associates for their generosity in sharing their data for research use. We thank all of our colleagues in the Google Brain Cambridge office for creating and maintaining such a supportive and stimulating environment. 

\bibliographystyle{unsrt}
\bibliography{bibliography}

\begin{thebibliography}{10}

\bibitem{Wu2018-yg}
Zhenqin Wu, Bharath Ramsundar, Evan~N Feinberg, Joseph Gomes, Caleb Geniesse,
  Aneesh~S Pappu, Karl Leswing, and Vijay Pande.
\newblock {MoleculeNet}: a benchmark for molecular machine learning.
\newblock {\em Chemical Science}, 9(2):513--530, 2018.

\bibitem{Mitchell2014-aw}
John B~O Mitchell.
\newblock Machine learning methods in chemoinformatics.
\newblock {\em Wiley Interdiscip. Rev. Comput. Mol. Sci.}, 4(5):468--481,
  September 2014.

\bibitem{Scarselli2009-nn}
Franco Scarselli, Marco Gori, Ah~Chung Tsoi, Markus Hagenbuchner, and Gabriele
  Monfardini.
\newblock Computational capabilities of graph neural networks.
\newblock {\em IEEE Trans. Neural Netw.}, 20(1):81--102, January 2009.

\bibitem{Gilmer2017-oq}
Justin Gilmer, Samuel~S Schoenholz, Patrick~F Riley, Oriol Vinyals, and
  George~E Dahl.
\newblock Neural message passing for quantum chemistry.
\newblock April 2017.

\bibitem{Rossiter1996-nc}
Karen~J Rossiter.
\newblock Structure-odor relationships.
\newblock {\em Chem. Rev.}, 96(8):3201--3240, 1996.

\bibitem{Su2009-bv}
Chih-Ying Su, Karen Menuz, and John~R Carlson.
\newblock Olfactory perception: receptors, cells, and circuits.
\newblock {\em Cell}, 139(1):45--59, October 2009.

\bibitem{McGann2017-ir}
John~P McGann.
\newblock Poor human olfaction is a 19th-century myth.
\newblock {\em Science}, 356(6338):eaam7263, 2017.

\bibitem{Yamins2016-nf}
Daniel L~K Yamins and James~J DiCarlo.
\newblock Using goal-driven deep learning models to understand sensory cortex.
\newblock {\em Nature Neuroscience}, 19(3):356--365, 2016.

\bibitem{Duvenaud2015-ye}
David~K Duvenaud, Dougal Maclaurin, Jorge Iparraguirre, Rafael
  G{\'o}mez-Bombarell, Timothy Hirzel, Al{\'a}n Aspuru-Guzik, and Ryan~P Adams.
\newblock {Convolutional Networks on Graphs for Learning Molecular
  Fingerprints}.
\newblock In {\em Advances in Neural Information Processing Systems}, pages
  2215--2223, 2015.

\bibitem{Altae-Tran2017-yz}
Han Altae-Tran, Bharath Ramsundar, Aneesh~S Pappu, and Vijay Pande.
\newblock Low data drug discovery with {One-Shot} learning.
\newblock {\em ACS Cent Sci}, 3(4):283--293, April 2017.

\bibitem{Fare2018-br}
Clyde Fare, Lukas Turcani, and Edward~O Pyzer-Knapp.
\newblock Powerful, transferable representations for molecules through
  intelligent task selection in deep multitask networks.
\newblock September 2018.

\bibitem{Sell2019-ti}
Charles Sell.
\newblock {\em Perfume in the Bible}.
\newblock Royal Society of Chemistry, July 2019.

\bibitem{Amoore1964-ca}
J~E Amoore, J~W Johnston, Jr, and M~Rubin.
\newblock {THE} {STEROCHEMICAL} {THEORY} {OF} {ODOR}.
\newblock {\em Sci. Am.}, 210:42--49, February 1964.

\bibitem{Dyson1937-ym}
G~Malcolm Dyson.
\newblock Raman effect and the concept of odour.
\newblock {\em Perfum. Essent. Oil Rec}, 28:13, 1937.

\bibitem{schiffman1974physicochemical}
Susan~S Schiffman.
\newblock Physicochemical correlates of olfactory quality.
\newblock {\em Science}, pages 112--117, 1974.

\bibitem{Chastrette1996-rm}
M~Chastrette, D~Cretin, and C~el~A{\"\i}di.
\newblock Structure-odor relationships: using neural networks in the estimation
  of camphoraceous or fruity odors and olfactory thresholds of aliphatic
  alcohols.
\newblock {\em J. Chem. Inf. Comput. Sci.}, 36(1):108--113, January 1996.

\bibitem{Sigma-Aldrich_Corporation2011-yk}
{Sigma-Aldrich Corporation}.
\newblock {\em Aldrich Chemistry 2012-2014: Handbook of Fine Chemicals}.
\newblock 2011.

\bibitem{Dravnieks1985-ek}
Andrew Dravnieks and {ASTM Committee E-18 on Sensory Evaluation of Materials
  and Products. Section E-18.04.12 on Odor Profiling}.
\newblock {\em Atlas of odor character profiles}.
\newblock Astm Intl, 1985.

\bibitem{Arctander1969-uf}
Steffen Arctander.
\newblock {\em Perfume and flavor chemicals:(aroma chemicals)}, volume~2.
\newblock Allured Publishing Corporation, 1969.

\bibitem{Sell2006-ag}
C~S Sell.
\newblock On the unpredictability of odor.
\newblock {\em Angewandte Chemie International Edition}, 45(38):6254--6261,
  2006.

\bibitem{Stumpfe2012-dc}
Dagmar Stumpfe and J{\"u}rgen Bajorath.
\newblock Exploring activity cliffs in medicinal chemistry.
\newblock {\em J. Med. Chem.}, 55(7):2932--2942, April 2012.

\bibitem{Keller2017-vs}
Andreas Keller, Richard~C Gerkin, Yuanfang Guan, Amit Dhurandhar, Gabor Turu,
  Bence Szalai, Joel~D Mainland, Yusuke Ihara, Chung~Wen Yu, Russ Wolfinger,
  Celine Vens, Leander Schietgat, Kurt De~Grave, Raquel Norel, {DREAM Olfaction
  Prediction Consortium}, Gustavo Stolovitzky, Guillermo~A Cecchi, Leslie~B
  Vosshall, and Pablo Meyer.
\newblock Predicting human olfactory perception from chemical features of odor
  molecules.
\newblock {\em Science}, 355(6327):820--826, February 2017.

\bibitem{Mauri2006-sp}
Andrea Mauri, Viviana Consonni, Manuela Pavan, and Roberto Todeschini.
\newblock Dragon software: An easy approach to molecular descriptor
  calculations.
\newblock {\em Match}, 56(2):237--248, 2006.

\bibitem{Rogers2010-uj}
David Rogers and Mathew Hahn.
\newblock Extended-connectivity fingerprints.
\newblock {\em J. Chem. Inf. Model.}, 50(5):742--754, May 2010.

\bibitem{Tran2018-nn}
Ngoc Tran, Daniel Kepple, Sergey~A Shuvaev, and Alexei~A Koulakov.
\newblock {DeepNose}: Using artificial neural networks to represent the space
  of odorants.
\newblock November 2018.

\bibitem{Nozaki2018-yg}
Yuji Nozaki and Takamichi Nakamoto.
\newblock Predictive modeling for odor character of a chemical using machine
  learning combined with natural language processing.
\newblock {\em PLoS One}, 13(6):e0198475, June 2018.

\bibitem{Gutierrez2018-ob}
E~Dar{\'\i}o Guti{\'e}rrez, Amit Dhurandhar, Andreas Keller, Pablo Meyer, and
  Guillermo~A Cecchi.
\newblock Predicting natural language descriptions of mono-molecular odorants.
\newblock {\em Nat. Commun.}, 9(1):4979, November 2018.

\bibitem{Ohloff2012-bg}
G{\"u}nther Ohloff, Wilhelm Pickenhagen, and Philip Kraft.
\newblock {\em Scent and Chemistry}.
\newblock Wiley, January 2012.

\bibitem{Svetnik2003-ht}
Vladimir Svetnik, Andy Liaw, Christopher Tong, J~Christopher Culberson,
  Robert~P Sheridan, and Bradley~P Feuston.
\newblock Random forest: a classification and regression tool for compound
  classification and {QSAR} modeling.
\newblock {\em J. Chem. Inf. Comput. Sci.}, 43(6):1947--1958, November 2003.

\bibitem{Moriwaki2018-hv}
Hirotomo Moriwaki, Yu-Shi Tian, Norihito Kawashita, and Tatsuya Takagi.
\newblock Mordred: a molecular descriptor calculator.
\newblock {\em J. Cheminform.}, 10(1):4, February 2018.

\bibitem{Maggiora2014-va}
Gerald Maggiora, Martin Vogt, Dagmar Stumpfe, and J{\"u}rgen Bajorath.
\newblock Molecular similarity in medicinal chemistry.
\newblock {\em J. Med. Chem.}, 57(8):3186--3204, April 2014.

\bibitem{Rogers2010-wo}
David Rogers and Mathew Hahn.
\newblock Extended-connectivity fingerprints.
\newblock {\em J. Chem. Inf. Model.}, 50(5):742--754, May 2010.

\bibitem{Randic1999-fn}
Milan Randi{\'c} and Subhash~C Basak.
\newblock Optimal molecular descriptors based on weighted path numbers.
\newblock {\em J. Chem. Inf. Comput. Sci.}, 39(2):261--266, March 1999.

\bibitem{rdkit}
{{RDK}it}: Open-source cheminformatics.
\newblock \url{http://www.rdkit.org}.

\bibitem{Wu2019}
Zonghan Wu, Shirui Pan, Fengwen Chen, Guodong Long, Chengqi Zhang, and
  Philip~S. Yu.
\newblock A comprehensive survey on graph neural networks.
\newblock {\em CoRR}, abs/1901.00596, 2019.

\bibitem{abadi2016tensorflow}
Mart{\'\i}n Abadi, Paul Barham, Jianmin Chen, Zhifeng Chen, Andy Davis, Jeffrey
  Dean, Matthieu Devin, Sanjay Ghemawat, Geoffrey Irving, Michael Isard, et~al.
\newblock Tensorflow: A system for large-scale machine learning.
\newblock In {\em 12th $\{$USENIX$\}$ Symposium on Operating Systems Design and
  Implementation ($\{$OSDI$\}$ 16)}, pages 265--283, 2016.

\bibitem{Bengio2013-gf}
Yoshua Bengio, Aaron Courville, and Pascal Vincent.
\newblock Representation learning: a review and new perspectives.
\newblock {\em IEEE Trans. Pattern Anal. Mach. Intell.}, 35(8):1798--1828,
  August 2013.

\bibitem{donahue2014decaf}
Jeff Donahue, Yangqing Jia, Oriol Vinyals, Judy Hoffman, Ning Zhang, Eric
  Tzeng, and Trevor Darrell.
\newblock Decaf: A deep convolutional activation feature for generic visual
  recognition.
\newblock In {\em International conference on machine learning}, pages
  647--655, 2014.

\bibitem{sharif2014cnn}
Ali Sharif~Razavian, Hossein Azizpour, Josephine Sullivan, and Stefan Carlsson.
\newblock Cnn features off-the-shelf: an astounding baseline for recognition.
\newblock In {\em Proceedings of the IEEE conference on computer vision and
  pattern recognition workshops}, pages 806--813, 2014.

\bibitem{Devlin2018-rp}
Jacob Devlin, Ming-Wei Chang, Kenton Lee, and Kristina Toutanova.
\newblock {BERT}: Pre-training of deep bidirectional transformers for language
  understanding.
\newblock October 2018.

\bibitem{coenen2019visualizing}
Andy Coenen, Emily Reif, Ann Yuan, Been Kim, Adam Pearce, Fernanda Vi{\'e}gas,
  and Martin Wattenberg.
\newblock Visualizing and measuring the geometry of {BERT}.
\newblock {\em arXiv preprint arXiv:1906.02715}, 2019.

\bibitem{Gomez-Bombarelli2018-ar}
Rafael G{\'o}mez-Bombarelli, Jennifer~N Wei, David Duvenaud, Jos{\'e}~Miguel
  Hern{\'a}ndez-Lobato, Benjam{\'\i}n S{\'a}nchez-Lengeling, Dennis Sheberla,
  Jorge Aguilera-Iparraguirre, Timothy~D Hirzel, Ryan~P Adams, and Al{\'a}n
  Aspuru-Guzik.
\newblock Automatic chemical design using a {Data-Driven} continuous
  representation of molecules.
\newblock {\em ACS Cent Sci}, 4(2):268--276, February 2018.

\bibitem{Zhavoronkov2019-pw}
Alex Zhavoronkov, Yan~A Ivanenkov, Alex Aliper, Mark~S Veselov, Vladimir~A
  Aladinskiy, Anastasiya~V Aladinskaya, Victor~A Terentiev, Daniil~A
  Polykovskiy, Maksim~D Kuznetsov, Arip Asadulaev, Yury Volkov, Artem Zholus,
  Rim~R Shayakhmetov, Alexander Zhebrak, Lidiya~I Minaeva, Bogdan~A
  Zagribelnyy, Lennart~H Lee, Richard Soll, David Madge, Li~Xing, Tao Guo, and
  Al{\'a}n Aspuru-Guzik.
\newblock Deep learning enables rapid identification of potent {DDR1} kinase
  inhibitors.
\newblock {\em Nat. Biotechnol.}, 37(9):1038--1040, September 2019.

\bibitem{noauthor_undated-uu}
The good scents company - flavor, fragrance, food and cosmetics ingredients
  information.
\newblock \url{http://www.thegoodscentscompany.com/}.
\newblock Accessed: 2019-9-4.

\bibitem{Leffingwell2005-uv}
John~C Leffingwell.
\newblock Leffingwell \& associates, 2005.

\bibitem{Pan2010-jk}
S~J Pan and Q~Yang.
\newblock A survey on transfer learning.
\newblock {\em IEEE Trans. Knowl. Data Eng.}, 22(10):1345--1359, October 2010.

\bibitem{Szymanski2017-di}
Piotr Szyma{\'n}ski and Tomasz Kajdanowicz.
\newblock A network perspective on stratification of {Multi-Label} data.
\newblock April 2017.

\end{thebibliography}

\section*{Supporting Information}

\beginsupplement


\subsection*{Hyperparameter Tunning and GNN Architecture \label{si:hyper}}
We consider two types of GNNs: Message Passing Neural Networks (MPNN) \cite{Gilmer2017-oq} and Graph Convolution Networks (GCN) \cite{Duvenaud2015-ye}. With both variants, we utilize a shared trunk that consists of message passing layers, followed by a reduce-sum operation, followed by several fully connected layers.

For the GCN and MPNN, we optimized the hyperparameters of our model using 5-fold cross-validation in our training set of $\sim$4,000 molecules, and tuned $\sim$30 hyperparameters (including learning rate, momentum, architecture depth \& width, etc) using 500 trials of random search. Each model fit took less than 1 hour on a Tesla P100. We present results for the model with the highest mean AUROC on the cross-validation set. 

We found that MPNNs and GCNs perform similarly. Both MPNNs and GCNs significantly outperform all baseline models. Because MPNNs and GCNs perform similarly, and GCNs are architecturally simpler, the analysis of GNN results in this work are reported on the GCN model.

\begin{table}[!h]
 \begin{tabularx}{\linewidth}{ c | Y Y}
 & GCN & MPNN \\\midrule
 Message Passing Layers & concatenation message type, 4 layers of dim: [15,20,27,36], selu activation, max graph pooling & edge-conditioned matrix multiply message type, 5 layers of dim 43, GRU-update at each layer\\
 Readout & Global sum pooling with softmax, 175 dim, one per MP layer and summed & Global sum pooling with softmax, 197 dim, one per MP layer with residual connections and summed \\ 
 fully-connected neural net & 2-layers of dim [96, 63] with relu, batchnorm, dropout of 0.47 & 3-layers of dim 392 with relu, batchnorm, dropout of 0.12 and l1/l2 regularization\\
 Prediction & \multicolumn{2}{c}{Multi-headed sigmoid, 138 tasks} \\
 \multirow{2}{*}{Training} & \multicolumn{2}{c}{Weighted-cross entropy loss, optimized with Adam,}\\
 & \multicolumn{2}{c}{used learning rate decay with warm restarts, 300 epochs} \\
\end{tabularx}
\mycaption{Tuned GNN architectures}{Hyperparameter settings from GCN and MPNN architectures.\label{table:gnn_architectures}}

\end{table}

\begin{table}[h]
    \centering
    \resizebox{\columnwidth}{!}{
\begin{tabular}{ccccc}
    \toprule
    & AUROC & Precision & Recall & F1 \\\midrule
    MPNN & 0.890 [0.882, 0.898] & 0.379 [0.352, 0.399] & 0.387 [0.366, 0.408] & 0.362 [0.335, 0.375] \\      
    GCN & 0.894 [0.888, 0.902] & 0.379 [0.351, 0.398] & 0.390 [0.365, 0.412] & 0.360 [0.337, 0.372] \\
    \bottomrule
\end{tabular}}
\mycaption{GCN and MPNN performance}{AUROC, Precision, Recall, and F1 results for odor prediction tasks for GCN and MPNN models. There are no appreciable differences between the MPNN and GCN performance. In the main text, GNN model refers to the GCN model. \label{table:gcn_mpnn_performance}}
\end{table}

For our RF baseline methods, we tuned an exhaustive space of configurations of fingerprinting methods (bits, radius, counted/binary, RDKit/Morgan), and RF hyperparameters. The RDKit software was used to calculate all features \cite{rdkit}.

For the KNN baseline methods, we also tuned fingerprinting options along with the number of neighbors. This resulted in a binary RDKit fingerprint of 4096 bits with radius 6. The optimal $k=20$ was found with an elbow analysis over $k=3$ to 100 using the Jaccard distance. KNN predictions are weighted by distance.

Since our multi-label problem had highly unbalanced labels, we used second-order iterative stratification to build our train/test/validation splits \cite{Szymanski2017-di}. Iterativative stratification is an iterative procedure for stratified sampling that attempts to preserve many-order label ratios, prioritizing more unbalanced combinations. For second order, this means preserving ratios of pairs of labels in each split.

\subsection*{Confidence Intervals}

Confidence intervals were constructed by bootstrap resampling. We resampled the test dataset with replacement $n=1000$ times, and computed AUROC on each sample. The training set and model remained fixed. We report the [2.5, 97.5] percentile boundaries to construct a 95\% CI interval.



\subsection*{Table of Per-Descriptor Results}
AUROC and AUPRC performance results by descriptor for the GNN model and the Random Forest model with counting fingerprint features.

\label{tab:detailed-classification-results}
\begin{longtable}{lrrrr}
\toprule
  & \multicolumn{2}{l}{AUROC} & \multicolumn{2}{l}{AUPRC} \\
  &    GNN &     RF-cFP &    GNN &     RF-cFP \\
\midrule    
Alcoholic     &  0.961 &  0.960 &  0.796 &  0.532 \\
Aldehydic     &  0.961 &  0.923 &  0.327 &  0.320 \\
Alliaceous    &  0.967 &  0.897 &  0.286 &  0.281 \\
Almond        &  0.943 &  0.933 &  0.509 &  0.132 \\
Amber         &  0.931 &  0.911 &  0.240 &  0.210 \\
Animal        &  0.812 &  0.837 &  0.198 &  0.183 \\
Anisic        &  0.791 &  0.881 &  0.383 &  0.101 \\
Apple         &  0.917 &  0.878 &  0.446 &  0.371 \\
Apricot       &  0.930 &  0.821 &  0.220 &  0.157 \\
Aromatic      &  0.855 &  0.766 &  0.056 &  0.029 \\
Balsamic      &  0.910 &  0.873 &  0.539 &  0.553 \\
Banana        &  0.961 &  0.913 &  0.541 &  0.207 \\
Beefy         &  0.976 &  0.953 &  0.231 &  0.261 \\
Bergamot      &  0.959 &  0.959 &  0.568 &  0.337 \\
Berry         &  0.858 &  0.808 &  0.142 &  0.127 \\
Bitter        &  0.719 &  0.801 &  0.241 &  0.071 \\
Black currant &  0.984 &  0.861 &  0.473 &  0.328 \\
Brandy        &  0.982 &  0.867 &  0.514 &  0.034 \\
Burnt         &  0.903 &  0.853 &  0.245 &  0.184 \\
Buttery       &  0.881 &  0.900 &  0.224 &  0.171 \\
Cabbage       &  0.977 &  0.913 &  0.189 &  0.332 \\
Camphoreous   &  0.951 &  0.916 &  0.410 &  0.239 \\
Caramellic    &  0.904 &  0.865 &  0.474 &  0.322 \\
Cedar         &  0.969 &  0.945 &  0.235 &  0.160 \\
Celery        &  0.878 &  0.753 &  0.201 &  0.041 \\
Chamomile     &  0.956 &  0.896 &  0.564 &  0.627 \\
Cheesy        &  0.920 &  0.882 &  0.315 &  0.191 \\
Cherry        &  0.905 &  0.926 &  0.198 &  0.230 \\
Chocolate     &  0.925 &  0.787 &  0.188 &  0.063 \\
Cinnamon      &  0.880 &  0.840 &  0.403 &  0.574 \\
Citrus        &  0.918 &  0.897 &  0.517 &  0.441 \\
Clean         &  0.878 &  0.786 &  0.072 &  0.122 \\
Clove         &  0.947 &  0.939 &  0.322 &  0.698 \\
Cocoa         &  0.938 &  0.927 &  0.355 &  0.388 \\
Coconut       &  0.959 &  0.860 &  0.525 &  0.348 \\
Coffee        &  0.938 &  0.911 &  0.392 &  0.457 \\
Cognac        &  0.979 &  0.950 &  0.488 &  0.298 \\
Cooked        &  0.848 &  0.795 &  0.243 &  0.193 \\
Cooling       &  0.973 &  0.878 &  0.342 &  0.299 \\
Cortex        &  0.759 &  0.629 &  0.181 &  0.024 \\
Coumarinic    &  0.950 &  0.873 &  0.565 &  0.160 \\
Creamy        &  0.808 &  0.674 &  0.185 &  0.092 \\
Cucumber      &  0.983 &  0.976 &  0.464 &  0.345 \\
Dairy         &  0.884 &  0.742 &  0.123 &  0.098 \\
Dry           &  0.731 &  0.679 &  0.206 &  0.124 \\
Earthy        &  0.745 &  0.728 &  0.234 &  0.190 \\
Ethereal      &  0.915 &  0.852 &  0.577 &  0.424 \\
Fatty         &  0.898 &  0.871 &  0.595 &  0.546 \\
Fermented     &  0.895 &  0.776 &  0.555 &  0.251 \\
Fishy         &  0.915 &  0.845 &  0.527 &  0.469 \\
Floral        &  0.852 &  0.843 &  0.539 &  0.535 \\
Fresh         &  0.756 &  0.715 &  0.263 &  0.222 \\
Fruit skin    &  0.840 &  0.710 &  0.123 &  0.146 \\
Fruity        &  0.859 &  0.842 &  0.797 &  0.743 \\
Garlic        &  0.986 &  0.979 &  0.610 &  0.522 \\
Gassy         &  0.986 &  0.871 &  0.566 &  0.214 \\
Geranium      &  0.905 &  0.829 &  0.327 &  0.204 \\
Grape         &  0.953 &  0.944 &  0.454 &  0.304 \\
Grapefruit    &  0.929 &  0.873 &  0.315 &  0.387 \\
Grassy        &  0.845 &  0.834 &  0.300 &  0.308 \\
Green         &  0.818 &  0.764 &  0.668 &  0.583 \\
Hawthorn      &  0.942 &  0.906 &  0.241 &  0.060 \\
Hay           &  0.775 &  0.690 &  0.045 &  0.032 \\
Hazelnut      &  0.987 &  0.992 &  0.330 &  0.654 \\
Herbal        &  0.766 &  0.723 &  0.246 &  0.222 \\
Honey         &  0.872 &  0.836 &  0.444 &  0.321 \\
Hyacinth      &  0.943 &  0.880 &  0.268 &  0.236 \\
Jasmine       &  0.951 &  0.943 &  0.454 &  0.321 \\
Juicy         &  0.907 &  0.646 &  0.089 &  0.209 \\
Ketonic       &  0.966 &  0.886 &  0.590 &  0.586 \\
Lactonic      &  0.919 &  0.898 &  0.339 &  0.205 \\
Lavender      &  0.961 &  0.920 &  0.364 &  0.286 \\
Leafy         &  0.842 &  0.776 &  0.218 &  0.126 \\
Leathery      &  0.830 &  0.698 &  0.070 &  0.041 \\
Lemon         &  0.855 &  0.804 &  0.387 &  0.272 \\
Lily          &  0.937 &  0.971 &  0.217 &  0.392 \\
Malty         &  0.888 &  0.586 &  0.045 &  0.004 \\
Meaty         &  0.945 &  0.896 &  0.482 &  0.396 \\
Medicinal     &  0.906 &  0.969 &  0.496 &  0.411 \\
Melon         &  0.884 &  0.843 &  0.214 &  0.344 \\
Metallic      &  0.758 &  0.705 &  0.386 &  0.156 \\
Milky         &  0.849 &  0.825 &  0.133 &  0.231 \\
Mint          &  0.898 &  0.846 &  0.488 &  0.472 \\
Muguet        &  0.922 &  0.900 &  0.262 &  0.115 \\
Mushroom      &  0.910 &  0.889 &  0.476 &  0.374 \\
Musk          &  0.917 &  0.832 &  0.659 &  0.521 \\
Musty         &  0.774 &  0.731 &  0.114 &  0.169 \\
Natural       &  0.811 &  0.752 &  0.077 &  0.053 \\
Nutty         &  0.844 &  0.827 &  0.508 &  0.372 \\
Odorless      &  0.973 &  0.955 &  0.754 &  0.660 \\
Oily          &  0.833 &  0.801 &  0.375 &  0.331 \\
Onion         &  0.979 &  0.961 &  0.673 &  0.559 \\
Orange        &  0.933 &  0.901 &  0.201 &  0.172 \\
Orangeflower  &  0.973 &  0.815 &  0.643 &  0.363 \\
Orris         &  0.910 &  0.855 &  0.228 &  0.138 \\
Ozone         &  0.957 &  0.887 &  0.120 &  0.467 \\
Peach         &  0.839 &  0.810 &  0.189 &  0.180 \\
Pear          &  0.952 &  0.926 &  0.456 &  0.351 \\
Phenolic      &  0.949 &  0.948 &  0.679 &  0.487 \\
Pine          &  0.956 &  0.895 &  0.306 &  0.207 \\
Pineapple     &  0.954 &  0.939 &  0.541 &  0.501 \\
Plum          &  0.862 &  0.791 &  0.109 &  0.346 \\
Popcorn       &  0.977 &  0.992 &  0.316 &  0.266 \\
Potato        &  0.983 &  0.973 &  0.282 &  0.246 \\
Powdery       &  0.881 &  0.831 &  0.150 &  0.132 \\
Pungent       &  0.886 &  0.845 &  0.554 &  0.433 \\
Radish        &  0.923 &  0.830 &  0.451 &  0.056 \\
Raspberry     &  0.827 &  0.737 &  0.190 &  0.146 \\
Ripe          &  0.911 &  0.868 &  0.162 &  0.055 \\
Roasted       &  0.932 &  0.907 &  0.510 &  0.431 \\
Rose          &  0.920 &  0.877 &  0.477 &  0.457 \\
Rummy         &  0.812 &  0.807 &  0.186 &  0.086 \\
Sandalwood    &  0.963 &  0.968 &  0.376 &  0.615 \\
Savory        &  0.944 &  0.877 &  0.254 &  0.298 \\
Sharp         &  0.810 &  0.701 &  0.210 &  0.044 \\
Smokey        &  0.909 &  0.903 &  0.343 &  0.305 \\
Soapy         &  0.885 &  0.764 &  0.154 &  0.114 \\
Solvent       &  0.857 &  0.580 &  0.078 &  0.017 \\
Sour          &  0.839 &  0.718 &  0.259 &  0.039 \\
Spicy         &  0.813 &  0.784 &  0.432 &  0.384 \\
Strawberry    &  0.909 &  0.873 &  0.062 &  0.073 \\
Sulfurous     &  0.983 &  0.982 &  0.666 &  0.754 \\
Sweaty        &  0.914 &  0.773 &  0.083 &  0.052 \\
Sweet         &  0.747 &  0.706 &  0.523 &  0.455 \\
Tea           &  0.745 &  0.744 &  0.061 &  0.245 \\
Terpenic      &  0.995 &  0.892 &  0.629 &  0.305 \\
Tobacco       &  0.877 &  0.899 &  0.104 &  0.269 \\
Tomato        &  0.941 &  0.925 &  0.151 &  0.147 \\
Tropical      &  0.848 &  0.782 &  0.367 &  0.256 \\
Vanilla       &  0.985 &  0.970 &  0.733 &  0.678 \\
Vegetable     &  0.884 &  0.816 &  0.232 &  0.242 \\
Vetiver       &  0.907 &  0.912 &  0.090 &  0.097 \\
Violet        &  0.833 &  0.753 &  0.254 &  0.142 \\
Warm          &  0.815 &  0.696 &  0.117 &  0.093 \\
Waxy          &  0.902 &  0.887 &  0.417 &  0.524 \\
Weedy         &  0.748 &  0.524 &  0.021 &  0.023 \\
Winey         &  0.902 &  0.833 &  0.359 &  0.284 \\
Woody         &  0.873 &  0.859 &  0.593 &  0.478 \\
\bottomrule
\end{longtable}


\begin{table}[h]
    \centering
\begin{tabular}{cc}
    \toprule
    Embedding space / distance metric & Kendall $\tau$ \\
    \midrule
    GCN embeddings with Euclidean distance & 0.280 \\
    GCN embeddings with cosine distance & 0.235 \\
    Morgan bit-FP with Jaccard distance & 0.187 \\
    Morgan bit-FP with cosine distance & 0.180 \\
    \bottomrule
\end{tabular}
\mycaption{Kendall Tau coefficient of embedding spaces and distance metrics}{
To assess which distance metric best correlated with label distance, we computed the Kendall Tau coefficient for each embedding space and metric. The Kendall Tau coefficient can be thought of intuitively as the fraction of the time that two pairwise distances are ordered correctly, ranging from [-1, 1] for reverse-sorted to sorted.
}
\label{tab:pairwise_distance_odor_label_table}
\end{table}

\end{document}